\pdfoutput=1
\documentclass[11pt]{article}

\usepackage[preprint]{acl}

\usepackage{times}
\usepackage{latexsym}
\usepackage{enumitem}
\usepackage[T1]{fontenc}
\usepackage[utf8]{inputenc}

\usepackage{microtype}

\usepackage{inconsolata}

\usepackage{graphicx}

\usepackage{hyperref}
\usepackage{url}
\usepackage{booktabs}       % professional-quality tables
\usepackage{amsfonts}       % blackboard math symbols
\usepackage{nicefrac}       % compact symbols for 1/2, etc.
\usepackage{multirow}
\usepackage{amsmath}
\usepackage{colortbl}
\usepackage{algorithm}
\usepackage{algpseudocode}
\usepackage{arydshln}
\definecolor{LightCyan}{rgb}{0.93,0.95,1}
\newcolumntype{b}{>{\columncolor{LightCyan}}c}
\usepackage{minitoc} % Package for local table of contents

\title{Medical Graph RAG: Towards Safe Medical Large Language Model via Graph Retrieval-Augmented Generation}
\author{
 \textbf{Junde Wu\textsuperscript{1}},
 \textbf{Jiayuan Zhu\textsuperscript{1}},
 \textbf{Yunli Qi\textsuperscript{1}},
 \textbf{Jingkun Chen\textsuperscript{1}},
\\
 \textbf{Min Xu\textsuperscript{2}},
 \textbf{Filippo Menolascina\textsuperscript{3}},
 \textbf{Vicente Grau\textsuperscript{1}},
\\
\\
 \textsuperscript{1}University of Oxford,
 \textsuperscript{2}Carnegie Mellon University,
 \textsuperscript{3}The University of Edinburgh,
\\
 \small{
   \textbf{Correspondence:} \href{mailto:email@domain}{jundewu@ieee.org}
 }
}

\begin{document}
\maketitle
\begin{abstract}
We introduce a novel graph-based Retrieval-Augmented Generation (RAG) framework specifically designed for the medical domain, called \textbf{MedGraphRAG}, aimed at enhancing Large Language Model (LLM) capabilities for generating evidence-based medical responses, thereby improving safety and reliability when handling private medical data. Graph-based RAG (GraphRAG) leverages LLMs to organize RAG data into graphs, showing strong potential for gaining holistic insights from long-form documents. However, its standard implementation is overly complex for general use and lacks the ability to generate evidence-based responses, limiting its effectiveness in the medical field. To extend the capabilities of GraphRAG to the medical domain, we propose unique Triple Graph Construction and U-Retrieval techniques over it. In our graph construction, we create a triple-linked structure that connects user documents to credible medical sources and controlled vocabularies. In the retrieval process, we propose U-Retrieval which combines Top-down Precise Retrieval with Bottom-up Response Refinement to balance global context awareness with precise indexing. These effort enable both source information retrieval and comprehensive response generation. Our approach is validated
on 9 medical Q\&A benchmarks, 2 health fact-checking benchmarks, and one collected dataset testing long-form generation. The results show that MedGraphRAG consistently outperforms state-of-the-art models across all benchmarks, while also ensuring that responses include credible source documentation and definitions. Our code is released at: \url{https://github.com/MedicineToken/Medical-Graph-RAG}.
\end{abstract}

\addtocontents{toc}{\protect\setcounter{tocdepth}{-10}} % Disable TOC for main sections

\section{Introduction}
%llm and current challenges
The rapid advancement of large language models (LLMs), such as OpenAI’s GPT-4 \cite{openai_gpt-4_2024}, has accelerated research in natural language processing and driven numerous AI applications. However, these models still face significant challenges in specialized fields like medicine \cite{hadi2024evaluation, williams2024evaluating, xie2024preliminary}. The first challenge is that these domains rely on vast knowledge bases -principles and notions discovered and accumulated over thousands years; fitting such knowledge into the finite context window of current LLMs is a hopeless task. Supervised Fine-Tuning (SFT) provides an alternative to using the context window, but it is often prohibitively expensive or unfeasible due to the closed-source nature of most commercial models. Second, medicine is a specialized field that relies on a precise terminology system and numerous established truths, such as specific disease symptoms or drug side effects. In this domain, it is essential that LLMs do not distort, modify, or introduce creative elements into the data. Unfortunately, verifying the accuracy of responses in medicine is particularly challenging for non-expert users. Therefore, the ability to perform complex reasoning using large external datasets, while generating accurate and credible responses backed by verifiable sources, is crucial in medical applications of LLMs.

%introduce rag and graphrag
Retrieval-augmented generation (RAG) \cite{lewis_retrieval-augmented_2021} is a technique that answers user queries using specific and private datasets without requiring further training of the model. However, RAG struggles to synthesize new insights and underperforms in tasks requiring a holistic understanding across extensive documents. GraphRAG \cite{hu_grag_2024} has been recently introduced to overcome these limitations. GraphRAG constructs a knowledge graph from raw documents using an LLM, and retrieves knowledge from the graph to enhance responses. By representing clear conceptual relationships across the data, it significantly outperforms classic RAG, especially for complex reasoning \cite{hu_grag_2024}. However, its graph construction lacks a specific design to ensure response authentication and credibility, and its hierarchical community construction process is costly, as it is designed to handle various cases for general-purpose use. We find that specific engineering effort is required to apply it effectively in the medical domain.

%our main idea and method
In this paper, we introduce a novel graph-based RAG method for medical domain, which we refer to as Medical GraphRAG (MedGraphRAG). This technique enhances LLM performance in the medical domain by generating evidence-based responses and official medical term explanation, which not only increases their credibility but also significantly improves their overall quality. Our method builds on GraphRAG with a more sophisticated graph construction technique, called Triple Graph Construction, to generate evidence-based responses, and an efficient retrieval method, U-Retrieval, which improves response quality with few costs. In Triple Graph Construction, we design a mechanism to link user RAG data to credible medical papers and foundational medical dictionaries. This process generates triples \textit{[RAG data, source, definition]} to construct a comprehensive graph of user documents. It enhances LLM reasoning and ensures responses are traceable to their sources and definitions, guaranteeing reliability and explainability. We also developed a unique U-Retrieval strategy to respond to user queries. Instead of building costly graph communities, we streamline the process by summarizing each graph using predefined medical tags, then iteratively clustering similar graphs to form a multi-layer hierarchical tag structure, from broad to detailed tags. The LLM generates tags for the user query and indexes the most relevant graph based on tag similarity in a top-down approach, using it to formulate the initial response. Then it refines the response by progressively integrating back the higher-level tags in a bottom-up manner until the final answer is generated. This U-Retrieval technique strikes a balance between global context awareness and the retrieval efficiency.

To evaluate our MedGraphRAG method, we implemented it on several popular open-source and commercial LLMs, including GPT \cite{chatgpt}, Gemini\cite{team2023gemini} and LLaMA \cite{touvron_llama_2023}. The results evaluated across 9 medical Q\&A benchmarks show that MedGraphRAG yielding materially better results than classic RAG and GraphRAG. Our final results even surpasses many specifically trained LLMs on medical corpora, setting a new state-of-the-art (SOTA) across all benchmarks. To verify its evidence-based response capability, we quantitatively tested MedGraphRAG on 2 health fact-checking benchmarks and conducted a human evaluation by experienced clinicians. Both evaluations strongly support that our responses are more source-based and reliable than previous methods.

Our contributions are as follows:

1. We are the first to propose a specialized framework for applying graph-based RAG in the medical domain, which we named MedGraphRAG.
   
2. We have developed unique Triple Graph Construction and U-Retrieval methods that enable LLMs to efficiently generate evidence-based responses utilizing holistic RAG data.

3. MedGraphRAG outperforms other retrieval methods and extensively fine-tuned Medical LLMs across a wide range of medical Q\&A benchmarks, establishing the new SOTAs.

4. Validated by human evaluations, MedGraphRAG is able to deliver more understandable and evidence-based responses in the medical domain.

\begin{figure*}[t]
    \begin{center}
    %\framebox[4.0in]{$\;$}
    \includegraphics[width=0.95\textwidth]{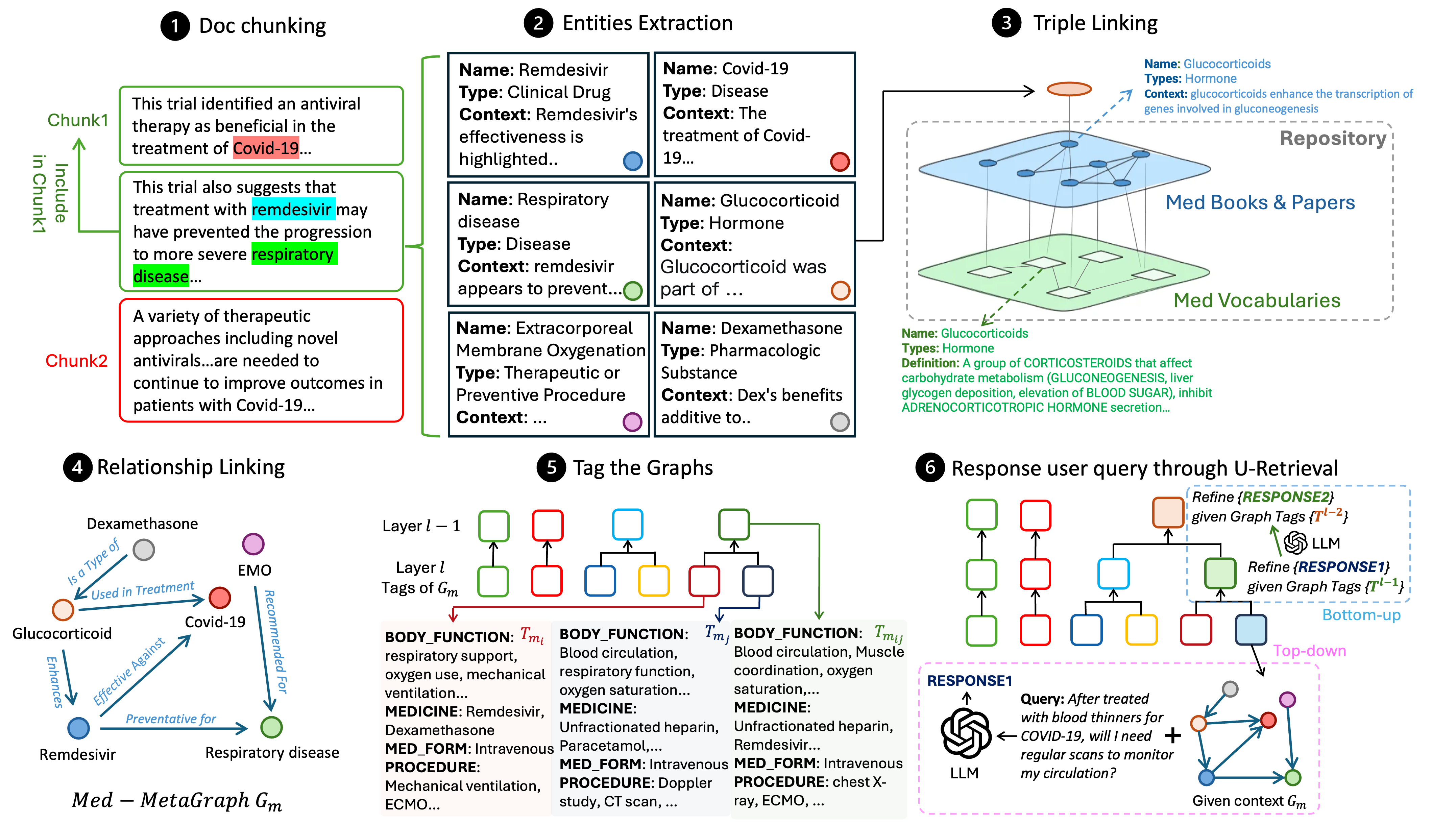}
    \end{center}
    \caption{The overall workflow of MedGraphRAG.}\label{fig1}
\end{figure*}

\section{Method}
\label{gen_inst}
The overall workflow of MedGraphRAG is shown in Fig. \ref{fig1}. We first construct the knowledge graphs based on the documents (Section \ref{sec:construct}), then organize and summarize the graphs for retrieval (Section \ref{sec:merge}), and finally retrieve data to response the user queries (Section \ref{sec:u-ret}).

\subsection{Medical Graph Construction}\label{sec:construct}
\subsubsection{Step1: Semantic Document Chunking}\label{sec:seg}
Large medical documents often contain multiple themes or diverse content. We first segment the document into data chunks that conform to the context limitations of LLMs. Traditional methods such as chunking based on token size or fixed characters typically fail to detect subtle shifts in topics accurately \cite{gao2023retrieval}. To enhance accuracy, we adopt a hybrid method combining character-based separation with topic-based semantic segmentation. Specifically, we first utilize line break symbols to isolate individual paragraphs \( P_i \) within the document $D = \{ P_1, P_2, \dots, P_{N_{p}} \}$. Then sequentially analyze the paragraphs, using the graph construction LLM $\mathcal{L}^{G}$ to determine if the current paragraph  \( P_j \)  should be included to the current chunk  \( H_j \) or start a new one, based on whether it shares the same topic as  \( H_j \). To reduce noise generated by sequential processing, we implement a sliding window managing $w$ paragraphs at a time, which is set as 5 in the paper. We slide the window in the first-in-first-out manner, maintaining focus on topic consistency, and use $\mathcal{L}^{G}$ token limitation $\text{Max}(\mathcal{L}^{G})$ as the hard threshold $\mathcal{T}$ of the chunks.  This document chunking process can be represented as
\[
\resizebox{0.9\hsize}{!}{$
H_j = \begin{cases}
H_{j-1} \cup \{ P_j \}, & \text{if } \mathcal{L}^{G}_{sem}(\{ P_{j-w+1}, \dots, P_j \}, \\
& H_{j-1}) = \textit{True} \text{ and } \\
& \sum_{P \in H_{j-1} \cup \{ P_j \}} \\
& \mathcal{T}(\phi (P)) \leq \text{Max}(\mathcal{L}^{G}); \\
\{ P_j \}, & \text{otherwise},
\end{cases}
$}
\]
where $\mathcal{L}^{G}_{sem}$ is $\mathcal{L}^{G}$ with semantic consistency prompt, $\phi$ is the tokenizer.

\subsubsection{Step2: Entities Extraction}\label{sec:ele}
We then extract entities from each chunk through $\mathcal{L}^{G}_{ent}$, which is $\mathcal{L}^{G}$ with entity extraction prompt. We prompt $\mathcal{L}^{G}$ to identify all relevant entities  $E = \{ e_1, e_2, \dots, e_{N^1_e} \}$ in each chunk $H$ and generate a structured output with \textit{name}, \textit{type}, and \textit{a description of the context}: $e = \{ na, ty, cx \}$, as the examples shown in the Step2 in Fig. \ref{fig1}. We set \textit{name} either be the text from the document or derivative term generated by $\mathcal{L}^{G}_{ent}$, \textit{type} one of the UMLS semantic types \cite{bodenreider2004unified}, and \textit{context} a few sentences generated by $\mathcal{L}^{G}_{ent}$ contextualized within the document. 

\subsubsection{Step3: Triple Linking}\label{sec:hie}
Medicine relies on precise terminology and established facts, making it essential for LLMs to produce responses grounded in established facts. To achieve this, we introduced Triple Graph Construction, linking user documents to credible sources and professional definitions. Specifically, we build repository graph (RepoGraph), which is intended to be fixed across different users, providing established sources and controlled vocabulary definitions for user RAG documents. We construct RepoGraph under user RAG graph with two layers: one based on medical papers/books and another based on medical dictionaries. We build the bottom layer of RepoGraph as UMLS \cite{bodenreider2004unified} graph, which consist comprehensive, well-defined medical vocabularies and their relationships . The upper layer of RepoGraph is constructed from medical textbooks and scholarly articles using the same graph construction method described here.

The entities of all three tiers of graphs are hierarchically linked through semantic relationships. Let us denoted entities extracted from RAG documents as $E^{1}$. We link them to entities extracted from medical books/papers, denoted as $E^{2}$, based on their relevance, which is determined by computing the cosine similarity between their content embeddings $\phi(C_e)$. The content of an entity $C_e$ is the concatenation of its \textit{name}, \textit{type}, and \textit{context}, represented as: $C_e = \text{Text}[\text{name: na; type: ty; context: cx}]$. This directed linking is annotated as \textit{the reference of}, indicating the reference relationship between entities in the two layers:
\scalebox{0.75}{
$R_{e^2}^{e^1} = \left\{ (e^{1}_i, The Reference Of, e^{2}_j) \ \middle| \ \frac{ \phi(C_{e^{1}_i}) \cdot \phi(C_{e^{2}_j)} }{ \| \phi(C_{e^{1}_i}) \| \, \| \phi(C_{e^{2}_j}) \| } \geq \delta_r \right\},
$
}
 where \( \delta_r \) is the pre-defined threshold. Entities $e^2 \in E^2$ are linked to $e^3 \in E^3$ through the same way with relationships annotated as \textit{the definition of} . Thus,  RAG entities are constructed as triples \textit{{[RAG entity, source, definition]}}.

Note that if we define the $i^{th}$-tier graph as $G^{i}$, representing a set of all entities and their relationships: $G^{i} = \{ E^i, R(E^i) \}$, where $R(E^i) = \{ R^{u}_{v} \mid u, v \in E^{i} \}$, we can then define the $k$ nearest triple neighbors of an entity $e_c$ as: $Tri^{\leq k}(e_{c}) = \bigcup_{i=0}^2 \left\{ e \in G^i \;\middle|\; \text{dist}(e_c, e) \leq k + i \right\}$, which includes all its $k$ nearest entities across three graph tiers.

\subsubsection{Step4: Relationship Linking}\label{sec:rela}
We then instruct $\mathcal{L}^{G}$ to identify the relationships among RAG entities in each chunk, which we noted as $e^1 \in E_m$. This relationship is a concise phrase generated by $\mathcal{L}^{G}$ based on the content of the entity $C_{e^1}$ and associated references $ \{C_{{e^2}}| R_{e^2}^{e1} = \text{the reference of} \}$. The identified relationships specify the source and target entities, provide a description of their relationship:
\scalebox{0.85}{
$
R_{e^1_i}^{e^1_j} = \left\{ (e^1_i, r_{ij}, e^1_j) \ \middle| \ \ r_{ij} = \mathcal{L}_{rel}^{G}(C_{e^1_i};C_{e_i^2} , C_{e^1_j};C_{e_j^2} ) \right\}, 
$
}
where $\mathcal{L}^{G}_{rel}$ is $\mathcal{L}^{G}$ with relationship identification and generation prompt. We show an example of relationship linking in the Step4 of Fig. \ref{fig1}. After performing this analysis, we have generated a directed graph for each data chunk, which is referred to as Meta-MedGraphs $G_{m} = \{E_{m}, R(E_m)\}$.

\subsection{Step5: Tag the Graphs}\label{sec:merge}
Organizing and summarizing the graphs in advance is intuitive and has proven to facilitate efficient retrieval \cite{hu_grag_2024}. However, unlike GraphRAG, we avoid constructing costly graph communities. We observe that, unlike general language content, medical text is often structured and can be summarized effectively using predefined tags. Motivated by this, we simply summarize each Meta-MedGraph $G_m$ with several predefined tags \( T \), and iteratively generate more abstract tag summaries for clusters of closely-related graphs. Specifically, \( \mathcal{L}^{G} \) first summarises the content of each Meta-MedGraph \( \{C_{e} \mid e \in  G_m\} \) given a set of given tags \( T \). The tags \( T \) consist of multiple categories, with the most important including \textit{Symptoms}, \textit{Patient History}, \textit{Body Functions}, and \textit{Medication}. The language model \( L^{G}_{\text{tag}} \) is then prompted with a system prompt template: 
\textit{Generate a structured summary from the provided medical content, strictly adhering to the following categories... \{Tag Name: Description of the tag\}...} . This process generates a structured tag-summary for each $G_m$, denoted as \( T_{m} \). 

We then apply a variant agglomerative hierarchical clustering method with dynamic thresholding based on tag similarity, to group the graphs and generate synthesized tag summaries. Initially, each graph begins as its own group. At each iteration, we compute the tag similarity between all pairs of clusters and dynamically set the threshold $\delta_t$ to merge the top 20\% most similar pairs. The graphs will be merged if all pairwise similarities within the group exceed $\delta_t$. Note that we don't really link the nodes across different graphs, but generate a synthesized tag-summary for each group. Specifically, we calculate the similarity of pairs by measuring the average cosine similarity of all their tag embeddings. Let \( \phi(t) \) denote the embedding of a tag \( t \in T_{m} \). Taking two Meta-MedGraphs $G_{m_i}$  and $G_{m_j}$ with tag sets \( T_{m_i} \) and \( T_{m_j} \) as an example, we generate the abstract tag summery $T_{m_{ij}}$ if their cosine similarity of tag embeddings \( \phi(t) \) and \( \phi(t') \) higher than the threshold $\delta_t$

\[
T_{m_{ij}} = \mathcal{L}_{mtag}^{G}(T_{m_i}, T_{m_j}), \quad \text{if } 
\]
\[
\frac{1}{|T_{m_i}| \cdot |T_{m_j}|} \sum\limits_{t \in T_{m_i}} \sum\limits_{t' \in T_{m_j}} \frac{ \phi(t)^\top \phi(t') }{ \| \phi(t) \| \, \| \phi(t') \| } \geq \delta_{t};
\]

where $\mathcal{L}_{mtag}^{G}$ is $L^{G}$ with tag-summary merging prompt. These newly merged tag-summary, along with those that remain unmerged, form a new layer of tags. As tag-summaries become less detailed at higher layers, there is a trade-off between precision and efficiency. In practice, we limit the process to 12 layers, as this is sufficient for most model variants. 

\subsection{Step6: U-Retrieval}\label{sec:u-ret}
After constructing the graph, we use response LLM $L^{R}$ efficiently retrieves information to respond to user queries through a strategy we called U-Retrieval. We begin by generating tag-summary on the user query $T_Q = \mathcal{L}^R(Q)$, and use these to identify the most relevant graph through a Top-down Precise Retrieval. Let's indicate the $j^{th}$ tags at layer $i$ summarised tag $T^{i}$ as $T^{i}[j]$, it starts from the top layer: $T^{0}$, progressively indexing down by selecting the most similar tag in each layer:
\[
T^{i+1}= \underset{T^{i}[j] \in T^{i}}{\arg\max} \ sim(T_{Q},T^{i}[j])
\]
until we reach the tag for the target Meta-MedGraph $G_{m_{t}}$. We then retrieve Top $N_u$ entities based on the embedding similarity between the query and the entity content:  $E_{r} = \left\{e \mid \text{Top} N_{u}(sim(\phi (Q), \phi (C_{e}))), e \in M_{t} \right\}$, and gather all their Top $k_{u}$  nearest triple neighbours $Tri^{\leq k_u}(e)$ as $E_{r}^{k_u} = \left\{e, Tri^{\leq k_u}(e), \mid e \in E_r\right\}$. 

By using all these entities and their relationships $G_r  = \{ E_{r}^{k_u}, R(E_{r}^{k_u})\}$, we prompt $\mathcal{L}^R$ to answer the question given the concatenated entity \textit{names} and \textit{relationships} in $G_r$: \textit{Given QUESTION: $\{Q\}$. GRAPH: $\{$$e_i [na]$+$R_{e_{i}}^{e^{j}}$+$e_{j} [na]$, ...$\}$. Answer the user question: QUESTION using the graph: GRAPH...} as $\mathcal{L}^R_{G_{r}}$. 

In the Bottom-up Response Refinement step, we then move back to the higher-level tag retrieved in the previous step $T^{i-1}$, in a bottom-up manner. We provide $\mathcal{L}^R$ \textit{QUESTION: $\{Q\}$}, \textit{LAST RESPONSE: ...}, and \textit{SUMMARY: $\{T^{i-1}\}$}, and ask it to \textit{Adjust the response:RESPONSE of the question: QUESTION using the updated information: SUMMARY}. $\mathcal{L}^R$ continues refining its responses until it reaches the target layer. In practice, we retrieve 4-6 layers depends on the baseline LLM. It ultimately generate a final response after scanning all indexed graphs along the trajectory. This method enables the LLM to gain a comprehensive overview by interacting with all relevant data in the graph, while remaining efficient by accessing less relevant data in summarized form. 

\section{Experiment}
\label{headings}

\begin{figure*}[h]
    \begin{center}
    %\framebox[4.0in]{$\;$}
    \includegraphics[width=\textwidth]{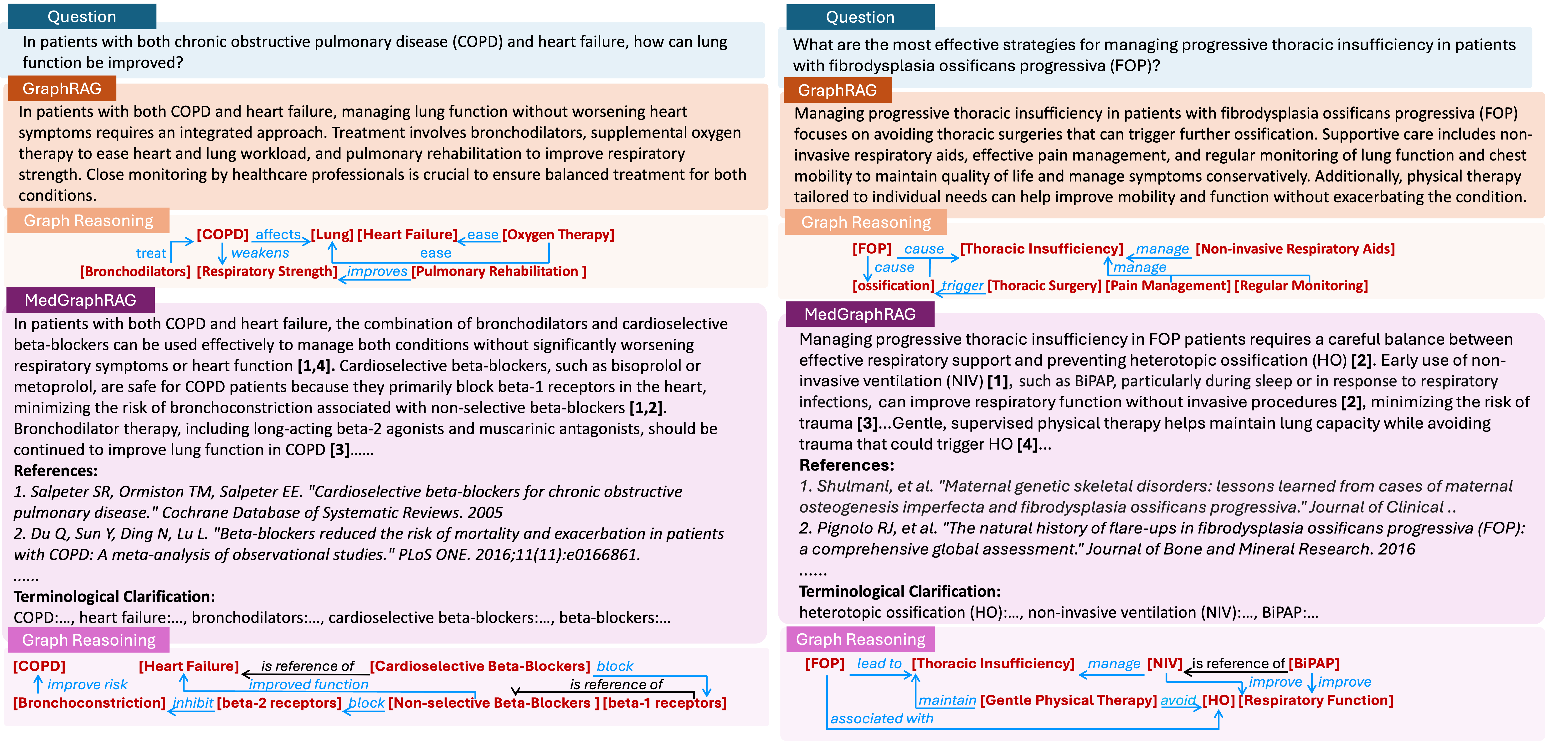}
    \end{center}
    \caption{Example responses from GraphRAG and MedGraphRAG, with abstracted graphs. MedGraphRAG provides more detailed explanations and more complex reasoning with evidences. }\label{fig2}
\end{figure*}

\subsection{Dataset}
\subsubsection{RAG data}
We anticipate that users will use frequently-updated private data as RAG data, such as patient electronic medical records.
% In our RAG data structure, we organize data into three tiers. The top tier contains user-specific, frequently-updated information, such as medical reports. 
Thus, we employ MIMIC-IV \cite{johnson2023mimic}, a publicly available electronic health record dataset, as RAG data. 
\subsubsection{Repository data}
We provide repository data to support LLM responses with credible sources and authoritative vocabulary definitions. We use MedC-K \cite{wu_pmc-llama_2023}, a corpus containing 4.8 million biomedical academic papers and 30,000 textbooks, along with all evidence publications from FakeHealth \cite{dai2020ginger} and PubHealth \cite{kotonya2020explainable}, as the upper repository data, and UMLS graph, which includes authoritative medical vocabularis and semantic relationships as the bottom repository data.

\subsubsection{Test Data}
Our test set are the test split of 9 multiple-choice biomedical datasets from the MultiMedQA suite, 2 fact verification dataset about public health, i.e., FakeHealth \cite{dai2020ginger} and PubHealth \cite{kotonya2020explainable}, and 1 test set we collected, called DiverseHealth. MultiMedQA includes MedQA \cite{jin2021disease}, MedMCQA \cite{pal2022medmcqa}, PubMedQA \cite{jin2019pubmedqa}, and MMLU clinic topics \cite{hendrycks2020measuring}. We also collected the DiverseHealth test set, focused on health equity, consisting of 50 real-world clinical questions that cover a wide range of topics, including rare diseases, minority health, comorbidities, drug use, alcohol, COVID-19, obesity, suicide, and chronic disease management. The dataset will be released alongside the paper. 

\subsection{Experiment Setting}
We compare different RAG methods across 6 language models as $\mathcal{L}^R$: Llama2 (13B, 70B), Llama3 (8B, 70B), Gemini-pro, and GPT-4. The Llama models were obtained from their official HuggingFace page. We used \textit{gemini-1.0-pro} for Gemini-pro and \textit{gpt-4-0613} for GPT-4. We primarily compare our approach with standard RAG implemented by LangChain\cite{langchain} and GraphRAG \cite{edge2024local} implemented by Microsoft Azure \cite{azure}. All retrieval methods are compared under same RAG data and test data.

We deploy $\mathcal{L}^G$ as \textit{Llama3-70B} to construct the graph. For text embeddings, we utilize OpenAI's \textit{text-embedding-3-large} model. Model comparison is performed using a 5-shot response ensemble \cite{li2024agentsneed}. MedGraphRAG used U-Retrieval with 4 levels on GPT-4, and 5 levels for the others. In the retrieval, we picked top 60 entities with their 16-hop neighbors. Unless otherwise noted, all thresholds are set as 0.5. We use the same query prompt for all models to generate responses.

\subsection{Results}
\subsubsection{Multi-Choice Evaluation}
\paragraph{Baselines with different retrievals}
First, we conducted experiments to evaluate retrieval methods on various LLM baselines, with the results shown in Table \ref{tab:main}. We compared MedGraphRAG against baselines without retrieval, standard RAG, and GraphRAG. Performance is measured by the accuracy of selecting the correct option. The results show that MedGraphRAG significantly enhances LLM performance on both health fact-checking and medical Q\&A benchmarks. Compared to baselines without retrieval, MedGraphRAG achieves an average improvement of nearly 10\% in fact-checking and 8\% in medical Q\&A. When compared to baselines using GraphRAG, it demonstrates an average improvement of around 8\% in fact-checking and 5\% in medical Q\&A. Notably, MedGraphRAG yields more pronounced improvements in smaller LLMs, such as $Llama2_{13B}$ and $Llama2_{8B}$. This suggests that MedGraphRAG effectively utilizes the models' own reasoning capabilities while providing them with additional knowledge beyond their parameters, serving as an external memory for information. 
\paragraph{Comparing with SOTA Medical LLMs}
When applied MedGraphRAG to larger models, like $Llama_{70B}$ or GPT, it resulted in new SOTA across all 11 datasets. This result also outperform intensively fine-tuning based medical large language models like Med-PaLM 2 \cite{singhal2023towards} and Med-Gemini \cite{saab2024capabilities}, establishing a new SOTA on the medical LLM leaderboard.

\begin{table*}[h]
\caption{Accuracy(\%) of LLMs using different retrieval methods. Columns with a blue background represent health fact-checking benchmarks, while the others correspond to medical Q\&A benchmarks. The best results are highlighted in bold.}
\resizebox{0.95\textwidth}{!}{
\begin{tabular}{cbbcccccccccc}
\hline
Model       & \begin{tabular}[c]{@{}c@{}}Fake\\ Health\end{tabular}        & \begin{tabular}[c]{@{}c@{}}Pub\\ Health\end{tabular} & MedQA & \begin{tabular}[c]{@{}c@{}}Med\\ MCQA\end{tabular}                                         & \begin{tabular}[c]{@{}c@{}}Pub\\ MedQA\end{tabular} &\begin{tabular}[c]{@{}c@{}}MMLU\\ Col-Med\end{tabular} &\begin{tabular}[c]{@{}c@{}}MMLU\\ Col-Bio\end{tabular} &\begin{tabular}[c]{@{}c@{}}MMLU\\ Pro-Med\end{tabular} &\begin{tabular}[c]{@{}c@{}}MMLU\\ Anatomy\end{tabular} &\begin{tabular}[c]{@{}c@{}}MMLU\\ Gene\end{tabular} &\begin{tabular}[c]{@{}c@{}}MMLU\\ Clinic\end{tabular}                                        \\ \hline
\multicolumn{11}{c}{\textit{Baselines without retrieval}} \\ 
% non retireval
Llama2-13B       & 53.8        & 49.4          & 42.7  & 37.4                                            & 68.0  & 60.7  & 69.4 & 60.3 & 52.6  & 66.0  & 63.8 \\

Llama2-70B        & 58.9       & 56.7          & 43.7  & 35.0                                            & 74.3  & 64.2  & 84.7 & 75.0 & 62.3  & 74.0  & 71.7                                           \\
Llama3-8B     & 51.1       & 53.2          & 59.8  & 57.3                                            & 75.2   & 61.9  & 78.5 & 70.2 & 68.9  & 83.0  & 74.7                                          \\
Llama3-70B     & 64.2          & 61.0          & 72.1  & 65.5                                            & 77.5    & 72.3  & 92.5 & 86.7 & 72.5  & 83.9  & 82.7      \\
Gemini-pro      & 60.6       & 63.7           & 59.0  & 54.8                                            & 69.8       & 69.2  & 88.0 & 77.7 & 66.7  & 75.8  & 76.7                                      \\
GPT-4          & 71.4        & 70.9           & 78.2  & 72.6                                            & 75.3      & 76.7 &95.3 &93.8 &81.3 &90.4 &86.2                                        \\ \cdashline{1-12}
% with rag
\multicolumn{11}{c}{\textit{Baselines with RAG}} \\ 
Llama2-13B      &  56.2       & 54.3          & 48.1  & 42.0                                            & 68.6  & 62.5  & 68.3 & 63.7 & 51.0  & 64.5  & 67.4 \\

Llama2-70B       & 64.6        & 63.2          & 56.2 & 49.8  & 75.2                                            & 69.6  & 85.8   & 77.4 & 63.0 & 75.8   & 73.3                                           \\
Llama3-8B          &60.5      & 59.6          & 64.3  & 58.2                                            & 76.0   & 68.6  & 84.9 & 73.2 & 72.1  & 85.2  & 77.8                                          \\
Llama3-70B       & 76.2       & 72.1          & 82.3  & 72.5                                            & 80.6    & 86.8  & 94.4 & 89.7 & 84.3  & 87.1  & 87.6                                         \\
Gemini-pro     &   72.5    & 68.4       & 64.5  & 57.3                                            & 76.9       & 79.0  & 91.3 & 86.4 & 79.5  & 80.4  & 83.9                                      \\
GPT-4        &78.6          & 77.3           & 88.1  & 76.3                                            & 77.6      & 81.2  & 95.5 & 94.3 & 83.1  & 92.9  & 93.1                                       \\ \cdashline{1-12}
% with graphrag
\multicolumn{11}{c}{\textit{Baselines with GraphRAG}} \\ 
Llama2-13B     &  58.7       & 57.5          & 52.3  & 44.6                                            & 72.8  & 64.1  & 73.0 & 64.6 & 52.1  & 66.2  & 67.9  \\

Llama2-70B      & 65.7        & 63.8          & 55.1 & 52.4  & 74.6                                            & 68.0  & 86.4   & 79.2 & 64.6 & 73.9   & 75.8                                           \\
Llama3-8B       &61.7      & 61.0          & 64.8  & 58.7                                            & 76.6   & 69.2  & 84.3 & 73.9 & 72.8  & 85.5  & 77.4                                          \\
Llama3-70B     & 77.7       & 74.5          & 84.1  & 73.2                                            & 81.2    & 87.4  & 94.8 & 89.8 & 85.2  & 87.9  & 88.5                                         \\
Gemini-pro      &   73.8    & 70.6       & 65.1  & 59.1                                            & 75.2       & 79.8  & 90.8 & 85.8 & 80.7  & 81.5  & 84.7                                      \\
GPT-4         &78.4          & 77.8           & 88.9  & 77.2                                            & 77.9      & 82.1  & 95.1 & 94.8 & 82.6  & 92.5  & 94.0                                       \\ \hline
% med graph rag
\multicolumn{11}{c}{\textit{\textbf{Baselines with MedGraphRAG}}} \\ 
Llama2-13B  &  64.1       & 61.2          & 65.5  & 51.4                                            & 73.2 & 68.4  & 76.5 & 67.2 & 56.0  & 67.3  & 69.5  \\
Llama2-70B  & 69.3        & 68.6          & 69.2  & 58.7                                            & 76.0    & 73.3  & 88.6   & 84.5 & 68.9 & 76.0   & 77.3                                           \\
Llama3-8B     & 79.9       & 77.6          & 74.2  &61.6 & 77.8  & 89.2  & 95.4 & 91.6 & 85.9  & 89.3  & 89.7                                         \\
Llama3-70B  & 81.2       & 79.2          & 88.4  & 79.1                                            & \textbf{83.8}     & 91.4  & 96.5 & 93.2 & 89.8  & 91.0  & 94.1                                         \\
Gemini-pro   &   79.2    & 76.4          & 71.8  & 62.0                                            & 76.2        & 86.3  & 92.9 & 89.7 & 85.0  & 87.1  & 89.3                                      \\
GPT-4   &\textbf{86.5}          & \textbf{83.4}           & \textbf{91.3}  & \textbf{81.5}                                            & 83.3    &\textbf{91.5} & \textbf{98.1} & \textbf{95.8} & \textbf{93.2} & \textbf{98.5} & \textbf{96.4}                                        \\ \hline
% Human (expert)     & -     & -            & 87.0  & 90.0                                            & 78.0    & 42.7  & 37.4 & 68.0 & 42.7  & 37.4  & 68.0     \\ \hline                                   
\end{tabular}}\label{tab:main}
\end{table*}

%-------------------------------------------------------------------------
\begin{figure}[h]
    \begin{center}
    %\framebox[4.0in]{$\;$}
    \includegraphics[scale=0.25]{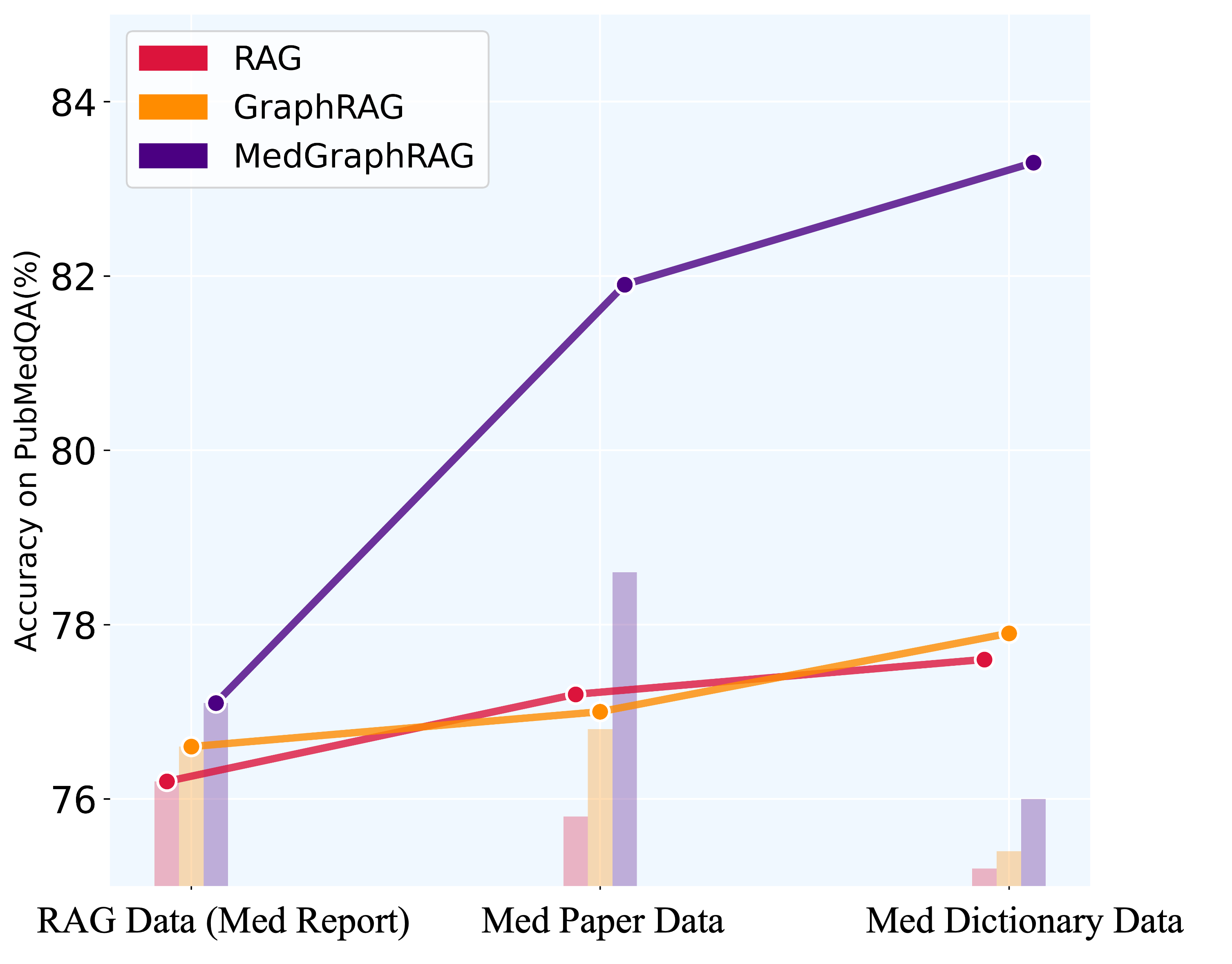}
    \end{center}
    \caption{Impact of Repository Data on RAG, GraphRAG, and MedGraphRAG with GPT-4. Line chart: performance with incremental data inclusion; bar chart: performance with individual data inclusion.}\label{fig:exp_ab2}
\end{figure}

\begin{table}[h]
\centering
\caption{Human evaluation on MedQA and DiverseHealth samples.}
%         \vspace{10pt}
        \resizebox{0.45\textwidth}{!}{
        \begin{tabular}{c|c|ccccc}
        \hline
Data                                                                              & Methods     & Pert.& Cor. & CP & CR & Und.\\ \hline
 \multirow{4}{*}{\begin{tabular}[c]{@{}c@{}}MultiMedQA\end{tabular}}& INLINE& 91& 88 & 80 & 74 &85\\
 & ATTR.FIRST& 93& 91&86 & 77 &93\\
& MIRAGE    & 95& 90& 84 & 75 &91\\
                                                                                  & MedGrapgRAG & \textbf{97}& \textbf{94}& \textbf{92} & \textbf{86}
                                                                                  & \textbf{95}\\  \hline
 \multirow{4}{*}{\begin{tabular}[c]{@{}c@{}}Diverse Health\end{tabular}}& INLINE& 95& 84& 78 & 71 &81\\
 & ATTR.FIRST& 96& 91& 81 & 78 &85\\ 
& MIRAGE    & 97& 89& 83 & 76 & 87\\
                                                                                  & MedGrapgRAG & \textbf{97}& \textbf{96}& \textbf{89} & \textbf{84}
                                                                                  & \textbf{93}\\ \hline

\end{tabular}}\label{tab:human}
\end{table}

\subsubsection{Long-form Generation Evaluation}
\paragraph{Human Evaluation} We conducted human evaluations of long-form model generation on the MultiMedQA and DiverseHealth benchmarks, comparing our method to SOTA models that generate citation-backed responses, including Inline Search in \cite{gao2023enabling}, ATTR-FIRST \cite{slobodkin2024attribute}, and MIRAGE \cite{qi2024model}. Our evaluation panel consisted of 7 certified clinicians and 5 laypersons to ensure feedback from both professional and general users. Raters completed a five-level rating survey for each model’s response, assessing responses across five dimensions: \textit{pertinence} (Pert.), \textit{correctness} (Cor.), \textit{citation precision} (CP), \textit{citation recall} (CR), and \textit{understandability} (Und.). As shown in Table \ref{tab:human}, MedGraphRAG consistently received higher ratings across all metrics. Notably, it showed a significant advantage in CP, CR and Und., indicating that its responses were more often backed by accurate sources and were easier to understand, even for laypersons, thanks to evidence-backed responses and clear explanations of complex medical terms.

\paragraph{Case Study}  
We compare the responses from GraphRAG and MedGraphRAG for a complex case involving patients with both chronic obstructive pulmonary disease (COPD) and heart failure (left plot). GraphRAG suggested general COPD treatments like bronchodilators and pulmonary rehabilitation but overlooked that certain bronchodilators may worsen heart failure symptoms. In contrast, MedGraphRAG provided a more comprehensive answer by recommending cardioselective beta-blockers—such as bisoprolol or metoprolol—that safely manage both conditions without adverse effects. As we can see form the graph abstracted, this superiority stems from MedGraphRAG's architecture, where entities are directly linked to key information in references, allowing retrieval of specific evidence. Conversely, GraphRAG struggles to retrieve specific information since its reference and user data are intertwined within the same layer of the graph, which leads to missing key information under the same number of nearest neighbors. And its retrieval based purely on graph summaries results in a lack of detailed insights.

\begin{table}[h]
\caption{An ablation study of MedGraphRAG, starting from GraphRAG, evaluated using accuracy (\%) on Q\&A datasets. }
\resizebox{0.45\textwidth}{!}{
\begin{tabular}{c|ccc}
                           & MedQA & PubMedQA & MedMCQA \\ \hline
GraphRAG                   & 88.9  & 77.9     & 77.2    \\
+Med-MetaGraph              & 90.4  & 78.5     & 78.1    \\
+Triple Graph Construction & 91.1  & 81.8     & 80.9    \\
+U-Retrieval (\textbf{MedGraphRAG})  & \textbf{91.3}  & \textbf{83.3}     & \textbf{81.5}   \\ \hline
\end{tabular}}\label{tab:ab}
\end{table}

\subsection{Analysis}
\subsubsection{Ablation Study based on GraphRAG}
We conducted a comprehensive ablation study to validate the effectiveness of our proposed modules, with the results presented in Table \ref{tab:ab}. Starting with GraphRAG \cite{hu_grag_2024} as the baseline, we incrementally incorporated our unique components, including the Med-MetaGraph implementation (entity structure, semantic document chunking, etc.), Triple Graph Construction, and U-Retrieval. Notably, both experiments were conducted on the same RAG dataset, eliminating data-related improvements. The results show a gradual performance improvement as more of our modules are added, with significant gains observed when replacing GraphRAG graph construction with our Triple Graph Construction.  Additionally, by replacing the summary-based retrieval\cite{edge_local_2024} in GraphRAG with our U-Retrieval method, we achieved further improvements, setting new state-of-the-art results across all three benchmarks.

\subsubsection{Which is important? Data or Method}
To assess the individual effects of external RAG data and retrieval technologies, we conducted experiments comparing retrieval methods: RAG, GraphRAG, and MedGraphRAG under two settings: (1) retrieving each tier of data separately (bar chart in Fig. \ref{fig:exp_ab2}), and (2) incrementally adding all three tiers (line chart in Fig. \ref{fig:exp_ab2}). The results show that both the data and the right retrieval method must work together to unlock the full potential. When retrieving data by standard RAG, Med-Paper data individually improves performance by less than 2\%, and Med-Dictionary data by less than 1\%. Accumulating three tier data also leads to mediocre improvements. GraphRAG shows improvement in retrieving individual data but has minimal gains when incrementally adding more data, likely due to superficiality from linking trivial entities, as discussed in the previous case study. In contrast, MedGraphRAG efficiently handles the additional data, using its hierarchical structure to clarify relationships and show strong improvements as more data is added. With MedGraphRAG, we see significant improvements of over 6\% and 8\% for Med-Paper and Med-Dictionary data, respectively, highlighting the importance of the retrieval method in maximizing the impact of the data.

\section{Related Work}
\subsection{LLM for Medicine} Large language models (LLMs) built on Transformer architectures have advanced rapidly, leading to specialized medical LLMs such as BioGPT \cite{luo2022biogpt}, PMC-LLaMA \cite{wu_pmc-llama_2023}, BioMedLM \cite{BioMedLM}, and Med-PaLM 2 \cite{singhal2023towards}. While many are fine-tuned by large organizations, recent research has focused on cost-efficient, non-fine-tuned approaches, primarily using prompt engineering \cite{saab2024capabilities,  wang2023prompt, savage2024diagnostic}. RAG, as another non-finetuning approach, is rarely explored for medical applications \cite{miao2024integrating, xiong2024benchmarking, long2024bailicai} and lacks support for evidence-based responses and term explanations required in clinical settings.

\subsection{Retrieval-augmented generation} RAG \cite{lewis_retrieval-augmented_2021} enables models to use specific datasets without additional training, improving response accuracy and reducing hallucinations \cite{guu_realm_2020}. RAG has shown strong results across various tasks, including generating responses with citations \cite{gao2023enabling, slobodkin2024attribute, qi2024model, nakano_webgpt_2021, bohnet_attributed_2022, gao_rarr_2023, gao_enabling_2023, schimanski_towards_2024, zhang2024longcite}. GraphRAG \cite{hu_grag_2024} further enhances complex reasoning by constructing knowledge graphs, but lacks specific design features for generating attributed responses, and its application in medical specialization remains limited.

\section{Conclusion}
MedGraphRAG improves the reliability of medical response generation with its graph-based RAG framework, using Triple Graph Construction and U-Retrieval to enhance evidence-based, context-aware responses. Its strong performance on benchmarks and human evaluations shows its ability to ensure accuracy in complex medical reasoning. Future work will focus on real-time data updates and validation on real-world clinical data.

\clearpage

\bibliography{acl_latex}

\end{document}